\RequirePackage{amsmath}

\documentclass[convention,peer-reviewed]{aesconf}

\graphicspath{{./}{figures/}}

\usepackage[utf8]{inputenc}

\usepackage{microtype}

\usepackage[numbers]{natbib}

\usepackage{booktabs}
\usepackage{color}
\usepackage{url}
\usepackage{amsfonts}
\usepackage{amssymb}
\usepackage{subfig}
\usepackage{float}
\usepackage[a4paper]{geometry}
\usepackage{graphicx} 
\usepackage{underscore}
\usepackage{booktabs, multirow} 
\usepackage{soul}
\usepackage[table]{xcolor} 
\usepackage{changepage,threeparttable} 

\title{Style Transfer for Non-differentiable Audio Effects}

\author[1]{Kieran Grant}

\affil[1]{University of Glasgow}

\correspondence{Kieran Grant}{kierang92@gmail.com}

\lastnames{Grant}

\shorttitle{Style Transfer for Non-differentiable Audio Effects}

\savebox{\AEStop}{%
	\begin{minipage}{\textwidth}%
		\rule{\textwidth}{1.5pt}\\%
		\\%
		\begin{minipage}[c][\iftoggle{convention}{3.2cm}{3.7cm}][t]{0\textwidth}%
			\includegraphics[width=20mm]{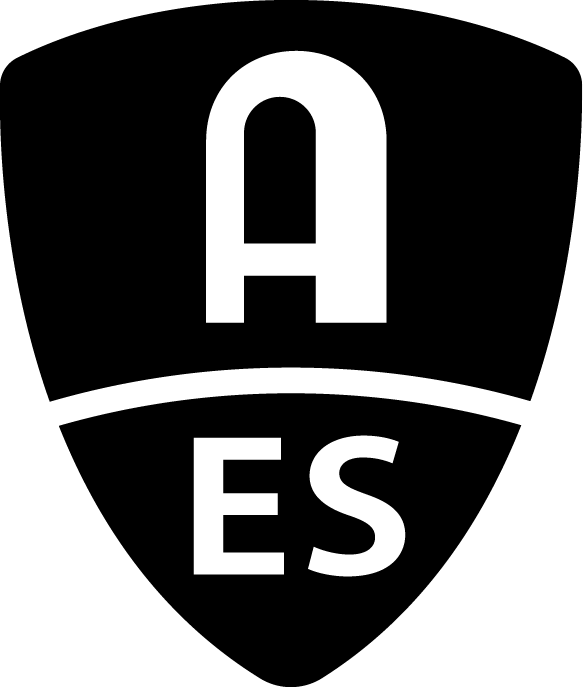}%
		\end{minipage}%
		\begin{minipage}{\textwidth}%
			\sffamily%
			\begin{center}%
				\LARGE Audio Engineering Society\\%
				\iftoggle{express_paper}{%
				\hspace{3mm}\fontsize{36}{38pt}\selectfont Convention Express\\Paper \AESExpressPaperNumber\\%
				}{%
				\iftoggle{convention}{%
				\fontsize{36}{38pt}\selectfont Convention Paper\\%
				}{%
				\fontsize{36}{38pt}\selectfont Conference Paper\\%
				}}%
				\vspace{0.2cm}%
				\large Presented at the \AESConferenceNumber \iftoggle{convention}{Convention\\}{\AESConferencePrefix Conference on\\}%
				\iftoggle{convention}{}{\AESConferenceTopic\\}%
				\AESConferenceDate\ifx\AESConferenceLocation\empty\else, \AESConferenceLocation\fi%
			\end{center}%
		\end{minipage}\\%
		\vspace{0.2cm}\\%
		\begin{minipage}{\textwidth}%
			\rmfamily\itshape\small	\AESLegalTextPrefix\ \AESLegalText%
		\end{minipage}\\%
		\\%
		\rule{\textwidth}{1.5pt}%
	\end{minipage}%
}

\begin{document}

\twocolumn[
\maketitle 

\begin{onecolabstract}
Digital audio effects are widely used by audio engineers to alter the acoustic and temporal qualities of audio data. However, these effects can have a large number of parameters which can make them difficult to learn for beginners and hamper creativity for professionals. Recently, there have been a number of efforts to employ progress in deep learning to acquire the low-level parameter configurations of audio effects by minimising an objective function between an input and reference track, commonly referred to as style transfer. However, current approaches use inflexible black-box techniques or require that the effects under consideration are implemented in an auto-differentiation framework. In this work, we propose a deep learning approach to audio production style matching which can be used with effects implemented in some of the most widely used frameworks, requiring only that the parameters under consideration have a continuous domain. Further, our method includes style matching for various classes of effects, many of which are difficult or impossible to be approximated closely using differentiable functions. We show that our audio embedding approach creates logical encodings of timbral information, which can be used for a number of downstream tasks. Further, we perform a listening test which demonstrates that our approach is able to convincingly style match a multi-band compressor effect.

\end{onecolabstract}
]


\begin{figure*}[!ht]
    \includegraphics[width=\textwidth]{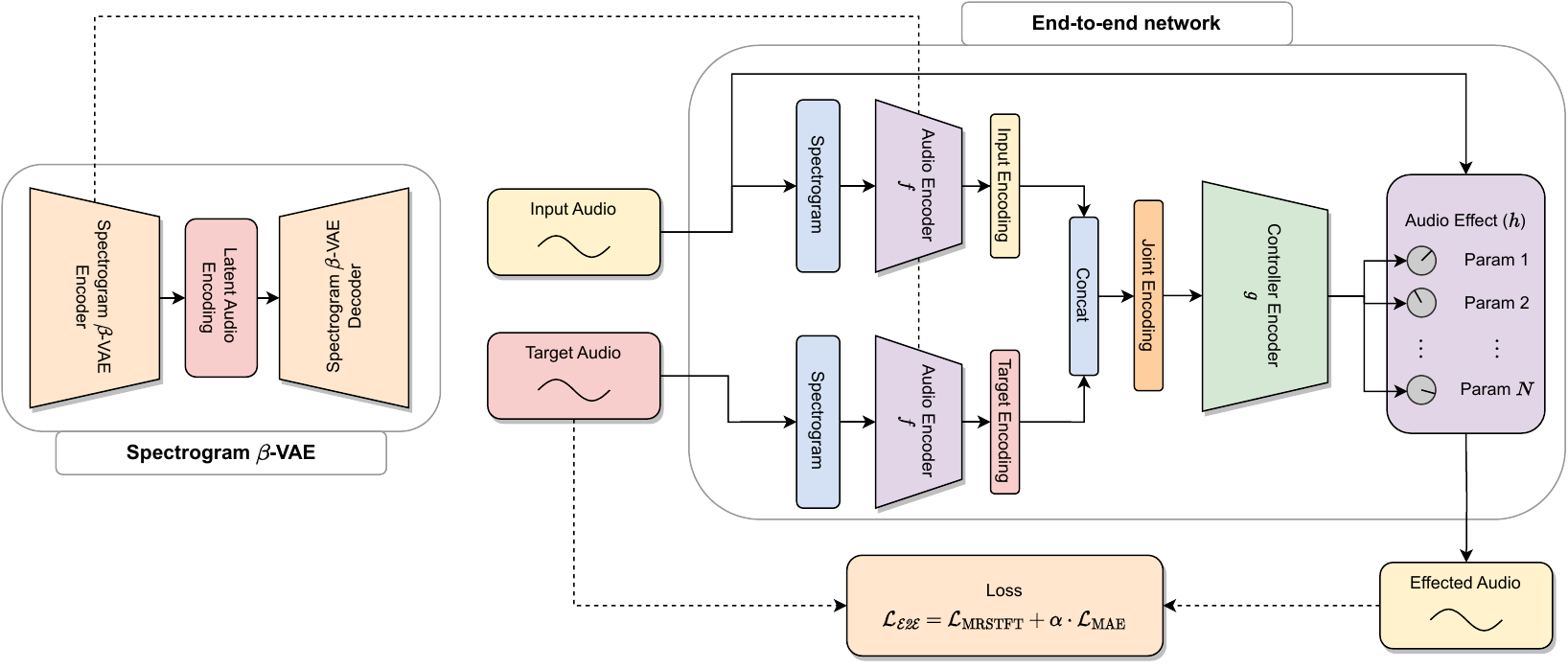}
    \caption{Architecture diagram of both the spectrogram $\beta$-VAE and the end-to-end network. The latter uses the pre-trained VAE audio encoder for creating embeddings of both the input and target audio.}
    \label{fig:archdiagram}
\end{figure*}

\section{Introduction}

Digital audio effects are a vital tool in a number of areas such as music production and sound design. However, these effects can have a large number of parameters which can be difficult to interpret and cause frustration for beginners. Similarly, experienced audio engineers can spend a large amount of time manually tweaking parameter settings in order to mix audio signals or for audio tasks such as mastering. 

The use of machine learning models to assist in controlling these effect parameters is becoming more common and can streamline the process of achieving the desired audio transformation. Recent solutions \citep{engel2020ddsp}\citep{steinmetz2022style} have explored the use of differentiable digital signal processing modules to ease the training of such machine learning models. However, these approaches generally either use black-box modelling techniques, which don’t allow for parameters to be adjusted after matching and lack interpretability \citep{vanhatalo2022review}, or require that the audio effects are implemented in an auto-differentiation framework \citep{nercessian2021lightweight}. 

In general, commercial audio effects are instead implemented in frameworks specifically designed for digital-signal processing tasks such as Steinberg’s Virtual Studio Technology (VST) format. Naturally, users will develop preferences for particular effect implementations which may make them hesitant to switch to a differentiable digital signal processing (DDSP) equivalent. This means that the practical application of these models which are only usable with auto-differentiable implementations of effects are limited. 

In this work, we address these gaps in the current literature by developing a method for audio style matching using generic (not necessarily differentiable) digital audio effects. To achieve this, we train a $\beta$-VAE \citep{higgins2016beta} network to capture audio features across a range of effect classes in a disentangled latent space. We then use the encoder section of this pre-trained $\beta$-VAE model in a Siamese network \citep{koch2015siamese} to learn a joint representation of both the input and reference audio. Finally, a simple feed-forward neural network is employed to transform these joint representations into parameter settings for the target effect.

In order to implement backpropagation during training, we employ a numerical gradient approximation method for the non-differentiable effect parameters \citep{ramirez2021differentiable}. After pre-training of the audio encoder model, the weights of the network up to the joint input-reference representation are frozen, allowing for the effect controller network to be retrained for any number of unseen audio effects. 

Our novel contributions in this work include the introduction of a pretrained, generalised audio encoder for downstream audio production tasks. By utilising this encoder, we demonstrate improved training stability and enhanced style matching performance compared to end-to-end training methods that rely on numerically estimated gradients for learning encoder weights. Additionally, we expand the range of effects that can be effectively incorporated in a style matching system. While our approach yields promising results for in-dataset effects, the generalisation to unseen effects remains an ongoing challenge. We provide an open-source implementation\footnote{Available from: \url{https://github.com/kieran-grant/nd-audio-style-transfer/}} of our approach, as well as listening examples\footnote{Available from: \url{https://nd-audio-style-transfer.streamlit.app/}}.



\section{Related Work}

Over the past decade, deep learning has found various applications in audio processing, such as text-to-speech synthesis \cite{arik2017deep}, genre/mood classification \cite{laurier2008multimodal}, and digital signal processing \cite{engel2020ddsp}\cite{vanhatalo2022review}. One application of deep learning to audio production is the development of black-box modeling techniques for analogue signal processing equipment. Some recent examples include the modeling of vintage guitar amplifiers \cite{vanhatalo2022review}\cite{damskagg2019deep} that contain non-linearities due to the distortion created by passing a high-gain signal through a vacuum tube, which are not easily replicated using traditional analytic methods. Nevertheless, such black-box approaches typically fix a parameter settings for a specific effect, requiring network retraining for each potential configuration.

An alternative strategy to black-box modeling of DAFX involves the utilisation of differentiable digital signal processing (DDSP) modules. Engel et al. \citep{engel2020ddsp} were among the first researchers to implement various classic DSP algorithms, such as filters, reverbs, and additive synthesisers, in an auto-differentiation framework. This approach enables the modules to be directly incorporated into neural networks and trained end-to-end using backpropagation. However, this DDSP approach presents a challenge, as each DSP module must be expressed as a differentiable transfer function. Furthermore, this approach does not readily extend to users who wish to utilise implementations of non-differentiable audio effects, which they may be familiar with, or which are more esoteric.

Inspired by the interpretability and flexibility of DDSP approaches, Ram{\'\i}rez et al. \citep{ramirez2021differentiable} have presented a method of end-to-end training using arbitrary DSP modules, i.e. those that are not necessarily differentiable. This approach relies on a numerical method for the estimation of gradients called Simultaneous Perturbation Stochastic Approximation (SPSA) \citep{spall1998overview} to allow for backpropagation. The approach is shown to be effective in a number of applications such as guitar tube-amplifier emulation, automatic music mastering and the removal of pops and breaths from recorded speech. However, the technique is yet to be used to allow arbitrary effects to be used for audio production style transfer. 

A common paradigm in audio style transfer is the use of Siamese networks \citep{koch2015siamese} which allow for meaningful representations of both input and reference audio to be learnt and combined for further downstream tasks such as parameter control. Typically, these networks use spectrogram representations of the audio and image encoders to learn dense representations. Sheng and Fazekas \citep{sheng2019feature} have utilised such a Siamese network to control a dynamic range compressor via a random forest controller network. In a separate study, Mimilakis et al. \citep{mimilakis2020one} proposed a method for controlling a parametric equaliser to match the vocal qualities of an input recording to a reference recording. The architecture used in this work is based on the Siamese architecture proposed by Steinmetz et al. \citep{steinmetz2022style} who implemented a differentiable equaliser and compressor for style transfer, while also comparing differentiation techniques such as neural proxies, SPSA and auto-differentiation for the task.



\section{Methodology}

\subsection{Model Architecture}

\subsubsection{Spectrogram $\beta$-VAE}

As part of the Siamese architecture of our model, we require a mapping from the audio domain to some latent encoding. To achieve this, we construct a $\beta$-VAE \citep{higgins2016beta} network with the objective of reconstructing a spectrogram representation of the input audio. Ideally, we aim for timbral qualities to be captured by this latent encoding which correspond directly to parameter settings for a wide range of generic audio effects. Hence, our latent space must be large enough to capture information about frequency and temporal content of the spectrogram, without being so large as to create many linearly dependent factors which would unnecessarily increase model complexity.

The spectrogram $\beta$-VAE itself comprises 4 convolutional layers, with batch normalisation and ReLU activations at each layer. The kernel (3), stride (2) and padding (1) sizes for each layer are shared, and the number of channels increase at each layer (8, 16, 32 and 32 respectively). The 32-channel image obtained from the convolutional network is then flattened and passed through two linear layers representing the $\mu$ and log-variance for our probabilistic sampling for a 128-dimensional latent space. The decoder is implemented as a mirror image of the encoder network using the relevant inverse function at each layer to create a reconstruction of the spectrogram from this latent space.

\subsubsection{Controller Network}

To map from these embeddings to parameter settings for a given effect module we employ a simple feed-forward controller network. This network takes the 256D concatenated input and reference encoding as inputs and maps these to the $P$ continuous parameters of the given effect. In our implementation we use a total of 4 linear hidden layers with 128, 128, 64 and 32 nodes respectively. At each layer we apply layer-normalisation followed by Leaky ReLU activation with a negative slope of $1\mathrm{e}{-3}$. At the final layer we use a sigmoid activation in order to map each of the $P$ network outputs in the range $(0, 1)$ which corresponds to the normalised values for each parameter.

The end-to-end network outputs the transformed audio signal, $\hat{y}$, by applying the predicted parameter configuration to the effect module and using this to transform the original input audio signal. As we make no assumption about the differentiability of the effect module under consideration, we employ numerical estimation techniques to calculate gradients during backpropagation. To achieve this, we adopt the approach proposed by Ram\'irez et al. \citep{ramirez2021differentiable}, which utilises the simultaneous permutation stochastic approximation (SPSA) method for efficient gradient estimation.

\begin{table}[ht]
\resizebox{\linewidth}{!}{
    \begin{tabular}{|c|c|}
    \hline
    \textbf{mda-vst Plugin Name}         & \textbf{Audio Effect Class}      \\ \hline
    Ambience                             & Reverb                   \\
    Combo                                & Amp simulator            \\
    Delay                                & Delay                    \\
    Dynamics                             & Compressor/limiter/gate  \\
    Overdrive                            & Soft distortion          \\
    RingMod                              & Ring modulation          \\ \hline
    \textit{Leslie}                      & \textit{Rotary speaker simulator} \\
    \textit{MultiBand}                   & \textit{Multi-band compressor}    \\
    \textit{Thru-Zero Flanger}           & \textit{Tape-flanging simulator}  \\ \hline
    \end{tabular}
}
    
\caption{mda-vst implementations and their respective general audio effect class. Effects in italics were not used during spectrogram $\beta$-VAE training.}
\label{tab:dafxnames}

\end{table}

\subsection{Dataset Generation}

\subsubsection{Digital Audio Effects}
For this work, we utilise a suite of legacy open-source audio plugins from mda-vst which are implemented using the Virtual Studio Technology 3 (VST3) framework, a format which is ubiquitous in real-world audio engineering. In order to interface with these plugins in a code-driven environment, we use Spotify’s Pedalboard library\footnote{Available from: \url{https://github.com/spotify/pedalboard}}, which allows us to process audio and update parameter settings directly. For each effect, we only consider adjustment of continuous parameters in order to be usable with our gradient estimation techniques described in the previous section. 

The mda-vst suite comprises over 30 audio plugins including software instruments and effects. For this work, we concentrate on a subset of widely used effect classes. These are given in Table \ref{tab:dafxnames}. 

\subsubsection{Audio}\label{sec:audiogen}

\begin{figure}[htb]
	\centering
	\includegraphics[scale=0.52]{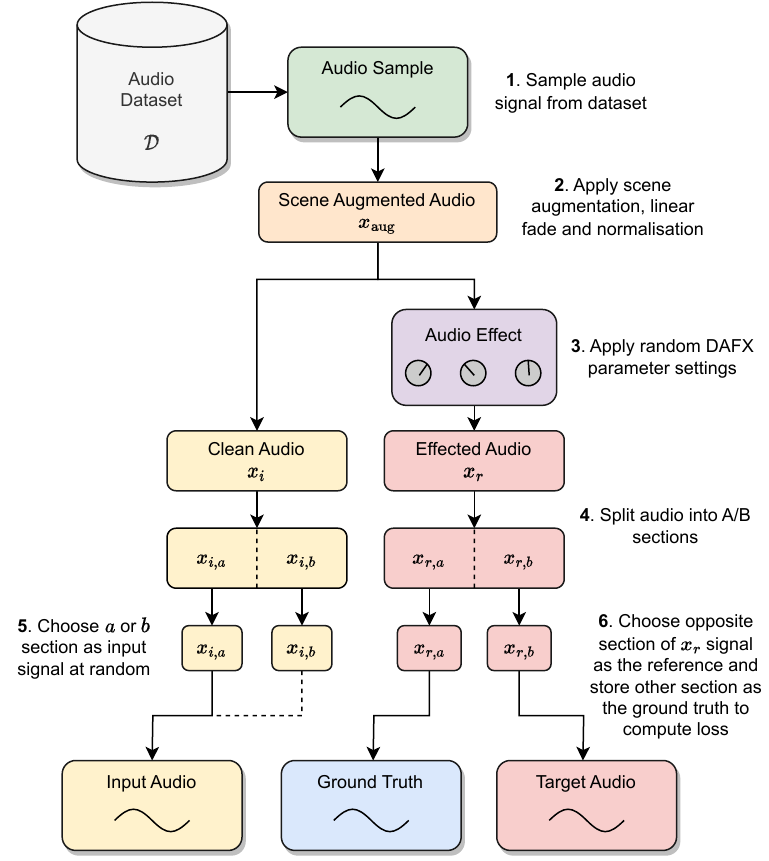}
	\caption{Dataset generation pipeline. Only the target audio segment is used during spectrogram $\beta$-VAE training, while paired input/target is used during end-to-end network training.}
    \label{fig:datagen}
\end{figure}

We adopt a data generation approach similar to Steinmetz et al. \citep{steinmetz2022style} and Mimilakis et al. \citep{mimilakis2020one} for self-supervised training without labeled or paired data. Figure \ref{fig:datagen} shows the main data generation pipeline.

Each training datapoint starts with sampling a random full-length audio recording from dataset $\mathcal{D}$. From a non-silent section of the source, a random patch of audio is taken as our audio sample. We apply scene augmentation, including pitch shifting and time-shifting, to create dataset variance. We then clone this augmented datapoint into two patches, $x_i$ and $x_r$. $x_i$ remains unaffected, while $x_r$ is processed by our chosen audio effect with a random parameter configuration $\theta$ and peak-normalized to -12dBFS to ensure similar levels between the input and reference, minimising the influence of volume in the style matching.

Matching audio production styles in real-world scenarios is challenging when the reference audio is from a different source than the input recording. To address this, we divide $x_i$ and $x_r$ into segments $a$ and $b$ of equal length. During model training, we randomly select either segment $a$ or $b$ as the input to the model, while the other segment serves as the reference. We then compute audio domain loss by comparing the held-out ground truth from the reference recording that corresponds to the same segment as the input recording.

\subsection{Model Training}

\subsubsection{Spectrogram $\beta$-VAE}
\label{sec:spectromethod}

We used a total of six different mda-vst effect modules for model training: Ambience, Delay, Dynamics, Overdrive, RingMod and Combo. We cycled through these effects such that a single effect was used at each training epoch. With this strategy we aimed to create a generalised encoder which can capture timbral characteristics which are universal across different effect classes. While the audio effect classes seen in training are not exhaustive of those used in real-world application, they provide a broad spectrum of effect types which create a variety of spectral, harmonic and temporal transformations to audio data.

To train the spectrogram $\beta$-VAE model, we applied a STFT to the target audio signal from Figure \ref{fig:datagen} with a length of $131,072$ samples at a sample rate of 24kHz (approximately 5 seconds of audio). The STFT was performed using 4096 FFT bins, with a Hann window size of 2048 and a hop length of 1024. The magnitude of the resulting complex-valued spectrogram was computed by taking the power of the absolute value of the spectrogram with an exponent of 0.3. The use of this exponential compression aided in enhancing the visual contrast of the image, thereby leading to a stronger distinction between the signal and ambient noise.

The normalised spectrogram was then used as the input for the spectrogram $\beta$-VAE. The weights of the network were randomly initialised, and we used a learning rate of $5\mathrm{e}{-4}$. We mitigated issues with vanishing KL-divergence loss and the resultant degrading effect on representations by employing cyclic KL-annealing \citep{fu2019cyclical} during training.

The final model was trained for 500 epochs with a training size of 2,500 and validation size of 250 examples per epoch. Training was performed using a single Nvidia 3070 RTX GPU and the weights which achieved the lowest overall validation loss were stored for downstream tasks.

\subsubsection{End-to-end Network}

For each of the nine audio effects under consideration, the Siamese network was trained end-to-end, with the weights for the audio encoder frozen from the earlier spectrogram $\beta$-VAE training. We followed previous work from Steinmetz et al. \citep{steinmetz2022style} and implemented an audio domain loss for our model training objective.  The overall loss is given by
\begin{equation}
	\mathcal{L_{\text{E2E}}} = \mathcal{L}_\text{MRSTFT} + \alpha \cdot \mathcal{L}_{\text{MAE}},
\end{equation}
Where $\mathcal{L}_{\text{MAE}}$ is the mean-absolute error between the predicted audio and the ground truth, $\alpha \geq 0$ is a weighting term, and $\mathcal{L}_{\text{MRSTFT}}$ is the Multi-Resolution STFT error which compares the predicted and target signals at multiple STFT resolutions \citep{steinmetz2020auraloss}. 

The end-to-end network was trained on CPU, with each effect trained for 30 epochs with 5,000 training examples per epoch and 500 validation examples per epoch. The learning rate was initialised to $1\mathrm{e}{-3}$ and a LR scheduler utilised which reduced the learning rate by a factor of 10 at 80\% and 95\% of training progress. Our $\alpha$ weighting of the $\mathcal{L}_\text{MAE}$ contribution was set to 100, and the $\epsilon$ value for SPSA was set to $1\mathrm{e}{-2}$.

\subsection{Datasets}
\label{sec:datasets}

We trained both the spectrogram $\beta$-VAE and end-to-end models using the Centre for Speech Technology Voice Cloning Toolkit (VCTK) corpus \citep{veaux2017cstr}. This dataset comprises speech data from 109 native English speakers with various accents, reading around 400 sentences each, mostly from newspaper passages. We reserve 11 speakers as a validation set.

For style matching evaluation, we utilise a subset of the Device and Produced Speech (DAPS) Dataset \citep{mysore2014can}, which includes studio-quality recordings from 20 English speakers, and the MusDB18 dataset \citep{rafii2017musdb18}, consisting of 150 full-track songs spanning different musical styles.



\section{Evaluation}
\label{sec:evaluation}

\subsection{Audio Encoder}
\label{sec:audioencoder}

\begin{table}[!htp]
    \centering
    \resizebox{\linewidth}{!}{
        \begin{tabular}{cccccc}\toprule
        &\multicolumn{2}{c}{\textbf{$\beta$-VAE}} &\multicolumn{2}{c}{\textbf{PCA}} \\\cmidrule{2-5}
        \textbf{Effect} &\textbf{Accuracy} &\textbf{F1-Score} &\textbf{Accuracy} &\textbf{F1-Score} \\\midrule
        \textbf{Ambience} &\cellcolor[HTML]{6464fa}\textcolor{white}{0.58} &\cellcolor[HTML]{5555fb}\textcolor{white}{0.65} &\cellcolor[HTML]{b5b5f6}0.25 &\cellcolor[HTML]{c2c2f5}0.19 \\
        \textbf{Combo} &\cellcolor[HTML]{0000ff}\textcolor{white}{1.00} &\cellcolor[HTML]{0000ff}\textcolor{white}{1.00} &\cellcolor[HTML]{c9c9f5}0.17 &\cellcolor[HTML]{cecef5}0.15 \\
        \textbf{Delay} &\cellcolor[HTML]{0606ff}\textcolor{white}{0.97} &\cellcolor[HTML]{0f0ffe}\textcolor{white}{0.94} &\cellcolor[HTML]{efeff3}0.01 &\cellcolor[HTML]{eeeef3}0.01 \\
        \textbf{Dynamics} &\cellcolor[HTML]{4444fc}\textcolor{white}{0.72} &\cellcolor[HTML]{4a4afb}\textcolor{white}{0.69} &\cellcolor[HTML]{dcdcf4}0.09 &\cellcolor[HTML]{dadaf4}0.10 \\
        \textbf{Overdrive} &\cellcolor[HTML]{0404ff}\textcolor{white}{0.98} &\cellcolor[HTML]{0a0afe}\textcolor{white}{0.96} &\cellcolor[HTML]{f0f0f3}0.00 &\cellcolor[HTML]{f0f0f3}0.00 \\
        \textbf{RingMod} &\cellcolor[HTML]{0000ff}\textcolor{white}{1.00} &\cellcolor[HTML]{0000ff}\textcolor{white}{1.00} &\cellcolor[HTML]{9595f7}\textcolor{white}{0.38} &\cellcolor[HTML]{b9b9f6}0.23 \\
        \bottomrule
        \end{tabular}
    }
\caption{Classification accuracy and F1-score on a holdout test set for a Random Forest (RF) classifier fitted separately to our spectrogram $\beta$-VAE embeddings and 128D PCA dimensionality reduction embeddings.}\label{tab:rfclassifer}
\end{table}

\subsubsection{Experiment Design}

To evaluate the effectiveness of our $\beta$-VAE model in separating audio effect classes, we utilised a random forest (RF) classifier to process the embeddings derived from the audio encoder mapping. We compare this non-linear transformation of audio data to a principal component analysis (PCA) approach to extract features from the original spectrograms. 

To standardise our dataset for analysis, we generated 6,000 data points (1,000 data points per 6 effect classes) with fixed parameter settings. These data points were transformed into 128D latent embeddings using both our trained encoder and PCA dimensionality reduction. We then randomly partitioned the embeddings into training (85\%) and test (15\%) sets, and subsequently trained the RF classifier separately on each of the embeddings. We report and compare the test accuracy and F1-scores of the RF classifier for both approaches. 

\begin{table*}[!htb]\centering
\resizebox{\textwidth}{!}{
    \begin{tabular}{ccccccccccccc}\toprule
        \multicolumn{2}{c}{\textbf{Overdrive}} &\multicolumn{2}{c}{\textbf{RingMod}} &\multicolumn{2}{c}{\textbf{Ambience}} &\multicolumn{2}{c}{\textbf{Combo}} &\multicolumn{2}{c}{\textbf{Delay}} &\multicolumn{2}{c}{\textbf{Dynamics}} \\\cmidrule{1-12}
        \textbf{Param} &\textbf{MMI} &\textbf{Param} &\textbf{MMI} &\textbf{Param} &\textbf{MMI} &\textbf{Param} &\textbf{MMI} &\textbf{Param} &\textbf{MMI} &\textbf{Param} &\textbf{MMI} \\\midrule
        muffle &\cellcolor[HTML]{0000ff}\textcolor{white}{2.34} &freq\_hz &\cellcolor[HTML]{a4a4f7}0.74 &mix &\cellcolor[HTML]{bbbbf5}0.53 &hpf\_freq &\cellcolor[HTML]{7272f9}\textcolor{white}{1.23} &fb\_mix &\cellcolor[HTML]{bebef5}0.49 &output\_db &\cellcolor[HTML]{e4e4f3}0.12 \\
        drive &\cellcolor[HTML]{8e8ef8}\textcolor{white}{0.96} &feedback &\cellcolor[HTML]{babaf6}0.54 &size\_m &\cellcolor[HTML]{d3d3f4}0.28 &hpf\_reso &\cellcolor[HTML]{c2c2f5}0.46 &l\_delay\_ms &\cellcolor[HTML]{cecef5}0.34 &gate\_thr\_db &\cellcolor[HTML]{e5e5f3}0.12 \\
        output\_db &\cellcolor[HTML]{f0f0f3}0.00 &fine\_hz &\cellcolor[HTML]{f0f0f3}0.00 &hf\_damp &\cellcolor[HTML]{e9e9f3}0.08 &drive\_s\_h &\cellcolor[HTML]{efeff3}0.02 &r\_delay &\cellcolor[HTML]{ebebf3}0.06 &mix &\cellcolor[HTML]{e7e7f3}0.10 \\
        & & & &output\_db &\cellcolor[HTML]{f0f0f3}0.00 &bias &\cellcolor[HTML]{f0f0f3}0.01 &feedback &\cellcolor[HTML]{eeeef3}0.03 &release\_ms &\cellcolor[HTML]{e9e9f3}0.08 \\
        & & & & & &output\_db &\cellcolor[HTML]{f0f0f3}0.00 &fb\_tone\_lo\_hi &\cellcolor[HTML]{f0f0f3}0.01 &gate\_rel\_ms &\cellcolor[HTML]{ededf3}0.04 \\
        & & & & & & & & & &limiter\_db &\cellcolor[HTML]{eeeef3}0.03 \\
        & & & & & & & & & &ratio &\cellcolor[HTML]{efeff3}0.02 \\
        & & & & & & & & & &thresh\_db &\cellcolor[HTML]{efeff3}0.02 \\
        & & & & & & & & & &gate\_att\_s &\cellcolor[HTML]{f0f0f3}0.01 \\
        & & & & & & & & & &attack\_s &\cellcolor[HTML]{f0f0f3}0.00 \\
        \\\cmidrule{1-12}
        \textbf{mean} &\cellcolor[HTML]{7f7ff9}\textcolor{white}{1.10} &\textbf{mean} &\cellcolor[HTML]{c5c5f5}0.43 &\textbf{mean} &\cellcolor[HTML]{dadaf4}0.22 &\textbf{mean} &\cellcolor[HTML]{cecef5}0.34 &\textbf{mean} &\cellcolor[HTML]{dedef4}0.19 &\textbf{mean} &\cellcolor[HTML]{ececf3}0.05 \\
    \bottomrule
    \end{tabular}
}
\caption{Maximum Mutual Information (MMI) calculated across 2D-projection of latent embeddings via Canonical Correlation Analysis (CCA) for each effect and parameter used during Spectrogram $\beta$-VAE training.}\label{tab:mmi}
\end{table*}

To evaluate the ability for the latent embeddings to capture timbral changes under different effect configurations we calculated the mutual information (MI) between each parameter for a given effect, and its relevant encodings. 

We first generated 10,000 parameter configurations for each audio effect and transformed a single audio datapoint under these configurations, computing the latent embedding obtained via our trained audio encoder. We performed Canonical Correlation Analysis (CCA) between the $10000 \times 128$ embedding matrix and $10000 \times P$ parameter settings matrix to obtain a $10000 \times 2$ reduced space. We then compared the MI for each parameter and axis in the reduced space in turn, and report the maximum MI value (MMI) achieved. 

Here, a larger MMI corresponds to a greater degree of variance in the embedding under changes to a particular parameter (thus capturing a specific and perceptible timbral change), while a MMI of 0 indicates that the latent embedding is not affected by changes to the relevant parameter. 

\subsubsection{Results and Discussion}

Results of the RF classifier experiment are presented in Table \ref{tab:rfclassifer}. Our audio encoder's embeddings consistently achieve higher accuracy and F1-scores across all effect classes compared to the PCA approach. This validating the use of a non-linear approach for embedding complex audio data. 

Table \ref{tab:mmi} displays the results of our experiment examining the audio embeddings' ability to capture timbral changes. Generally, we observe a strong correlation with changes in the audio signal's frequency content. For example, parameters like Overdrive's \texttt{muffle} and RingMod's \texttt{freq_hz} significantly impact the audio's harmonic content and achieve the highest MMI in their respective classes. Our audio and spectrogram normalisation have also successfully mitigated the influence of volume on embeddings. However, some parameter settings exhibit little correlation with changes in the audio's latent embedding. In particular, subtle compression and delay settings show little correlation to their embeddings.

\begin{table*}[ht]
\centering
\resizebox{\textwidth}{!}{
\begin{tabular}{ccccccccccc}
\hline
                                                        &                                           & \multicolumn{9}{c}{\textbf{Audio Effect}}                                                                                                                                                                                                                                                                                                                                      \\ \cline{3-11} 
\textbf{Dataset}                                        & \textbf{Audio Encoder}                    & \textbf{Ambience}                      & \textbf{Combo}                         & \textbf{Delay}                         & \textbf{Dynamics}                      & \textbf{Flanger}                       & \textbf{Leslie}                        & \textbf{MultiBand}                     & \textbf{Overdrive}                     & \textbf{RingMod}                       \\ \hline
\rowcolor[HTML]{EFEFEF} 
\cellcolor[HTML]{FFFFFF}                                & \textit{Baseline}                         & \textbf{0.443}                         & 1.449                                  & \textbf{1.053}                         & \textbf{0.613}                         & \textbf{0.665}                         & \textbf{0.497}                         & 0.629                                  & 1.138                                  & 1.777                                  \\
\cellcolor[HTML]{FFFFFF}                                & Untrained                                 & 0.600                                  & 1.525                                  & 1.218                                  & 0.803                                  & 0.806                                  & 0.641                                  & 0.608                                  & 0.794                                  & 1.712                                  \\
\rowcolor[HTML]{EFEFEF} 
\multirow{-3}{*}{\cellcolor[HTML]{FFFFFF}\textbf{VCTK}} & Pretrained                                   & 0.525                                  & \textbf{1.055}                         & 1.078                                  & 0.701                                  & 0.696                                  & 0.621                                  & \textbf{0.492}                         & \textbf{0.640}                         & \textbf{1.621}                         \\ \hline
                                                        & \textit{Baseline}                         & \textbf{0.514}                         & 1.496                                  & \textbf{1.025}                         & \textbf{0.595}                         & 0.753                                  & \textbf{0.519}                         & 0.662                                  & 1.245                                  & 1.858                                  \\
                                                        & \cellcolor[HTML]{EFEFEF}Untrained         & \cellcolor[HTML]{EFEFEF}0.650          & \cellcolor[HTML]{EFEFEF}1.534          & \cellcolor[HTML]{EFEFEF}1.152          & \cellcolor[HTML]{EFEFEF}0.799          & \cellcolor[HTML]{EFEFEF}0.880          & \cellcolor[HTML]{EFEFEF}0.657          & \cellcolor[HTML]{EFEFEF}0.602          & \cellcolor[HTML]{EFEFEF}0.882          & \cellcolor[HTML]{EFEFEF}1.988          \\
\multirow{-3}{*}{\textbf{DAPS}}                         & Pretrained                                   & 0.545                                  & \textbf{1.048}                         & 1.028                                  & 0.648                                  & \textbf{0.742}                         & 0.607                                  & \textbf{0.472}                         & \textbf{0.639}                         & \textbf{1.856}                         \\ \hline
                                                        & \cellcolor[HTML]{EFEFEF}\textit{Baseline} & \cellcolor[HTML]{EFEFEF}\textbf{0.454} & \cellcolor[HTML]{EFEFEF}1.601          & \cellcolor[HTML]{EFEFEF}\textbf{0.703} & \cellcolor[HTML]{EFEFEF}\textbf{0.482} & \cellcolor[HTML]{EFEFEF}\textbf{0.692} & \cellcolor[HTML]{EFEFEF}\textbf{0.620} & \cellcolor[HTML]{EFEFEF}0.964          & \cellcolor[HTML]{EFEFEF}1.534          & \cellcolor[HTML]{EFEFEF}\textbf{1.459} \\
                                                        & Untrained                                 & 0.646                                  & 1.518                                  & 0.905                                  & 1.062                                  & 0.920                                  & 0.791                                  & 0.888                                  & 1.101                                  & 2.598                                  \\
\multirow{-3}{*}{\textbf{MusDB18}}                      & \cellcolor[HTML]{EFEFEF}Pretrained           & \cellcolor[HTML]{EFEFEF}0.518          & \cellcolor[HTML]{EFEFEF}\textbf{1.040} & \cellcolor[HTML]{EFEFEF}0.765          & \cellcolor[HTML]{EFEFEF}0.554          & \cellcolor[HTML]{EFEFEF}0.748          & \cellcolor[HTML]{EFEFEF}0.727          & \cellcolor[HTML]{EFEFEF}\textbf{0.725} & \cellcolor[HTML]{EFEFEF}\textbf{0.797} & \cellcolor[HTML]{EFEFEF}2.410          \\ \hline
\end{tabular}
}
\caption{Multi-Resolution STFT (MRSTFT) loss across the nine effect classes and three datasets.  For each effect-dataset pair, we calculate mean loss across 5,000 examples. In this instance a lower score is better.}\label{tab:mrstft}
\end{table*}


\begin{table*}[ht]
\centering
\resizebox{\textwidth}{!}{
\begin{tabular}{ccccccccccc}
\hline
\rowcolor[HTML]{FFFFFF} 
\multicolumn{1}{l}{\cellcolor[HTML]{FFFFFF}}               & \multicolumn{1}{l}{\cellcolor[HTML]{FFFFFF}} & \multicolumn{9}{c}{\cellcolor[HTML]{FFFFFF}\textbf{Audio Effect}}                                                                                                         \\ \cline{3-11} 
\rowcolor[HTML]{FFFFFF} 
\textbf{Dataset}                                           & \textbf{Audio Encoder}                       & \textbf{Ambience} & \textbf{Combo} & \textbf{Delay} & \textbf{Dynamics} & \textbf{Flanger} & \textbf{Leslie} & \textbf{MultiBand} & \textbf{Overdrive} & \textbf{RingMod} \\ \hline
\rowcolor[HTML]{EFEFEF} 
\cellcolor[HTML]{FFFFFF}                                   & \textit{Baseline}                            & 2.621             & 2.600          & \textbf{1.817} & 4.174             & \textbf{2.418}   & \textbf{3.436}  & 4.358              & 3.073              & \textbf{1.537}   \\
\rowcolor[HTML]{FFFFFF} 
\cellcolor[HTML]{FFFFFF}                                   & Untrained                                    & 2.027             & \textbf{3.609} & 1.306          & 3.957             & 1.670            & 2.990           & 4.373              & 4.014              & 1.120            \\
\rowcolor[HTML]{EFEFEF} 
\multirow{-3}{*}{\cellcolor[HTML]{FFFFFF}\textbf{VCTK}}    & Pretrained                                   & \textbf{2.625}    & 3.403          & 1.611          & \textbf{4.207}    & 2.386            & 3.316           & \textbf{4.451}     & \textbf{4.224}     & 1.148            \\ \hline
\rowcolor[HTML]{FFFFFF} 
\cellcolor[HTML]{FFFFFF}                                   & \textit{Baseline}                            & \textbf{2.439}    & 2.359          & \textbf{1.736} & 4.167             & \textbf{2.285}   & \textbf{3.255}  & 4.385              & 2.927              & \textbf{1.458}   \\
\rowcolor[HTML]{EFEFEF} 
\cellcolor[HTML]{FFFFFF}                                   & Untrained                                    & 1.742             & \textbf{3.564} & 1.203          & 3.925             & 1.446            & 2.683           & 4.398              & 4.059              & 1.078            \\
\rowcolor[HTML]{FFFFFF} 
\multirow{-3}{*}{\cellcolor[HTML]{FFFFFF}\textbf{DAPS}}    & Pretrained                                   & 2.407             & 3.378          & 1.534          & \textbf{4.209}    & 2.039            & 3.133           & \textbf{4.486}     & \textbf{4.213}     & 1.097            \\ \hline
\rowcolor[HTML]{EFEFEF} 
\cellcolor[HTML]{FFFFFF}                                   & \textit{Baseline}                            & \textbf{2.862}    & 1.786          & \textbf{2.446} & 4.075             & \textbf{2.611}   & \textbf{2.605}  & 4.004              & 3.557              & \textbf{1.511}   \\
\rowcolor[HTML]{FFFFFF} 
\cellcolor[HTML]{FFFFFF}                                   & Untrained                                    & 1.805             & \textbf{3.736} & 1.475          & 3.707             & 1.450            & 2.117           & 4.134              & 4.040              & 1.088            \\
\rowcolor[HTML]{EFEFEF} 
\multirow{-3}{*}{\cellcolor[HTML]{FFFFFF}\textbf{MusDB18}} & Pretrained                                   & 2.780             & 3.689          & 2.324          & \textbf{4.177}    & 2.339            & 2.518           & \textbf{4.272}     & \textbf{4.221}     & 1.090            \\ \hline
\end{tabular}
}
\caption{Perceptual Evaluation of Speech Quality (PESQ) scores across the nine effect classes and three datasets. The output of the PESQ algorithm is a value in the range 1 (\textit{poor}) to 5 (\textit{excellent}) \citep{recommendation2007wideband}.}\label{tab:pesq}

\end{table*}

\subsection{End-to-End Network}
\label{sec:end2endeval}

\subsubsection{Experiment Design}

To assess the performance of our end-to-end network for the task of style transfer for unseen audio sources, we conducted a comparative analysis using the three datasets discussed in Section \ref{sec:datasets}. We used nine separately trained instances of our end-to-end network (one for each of the audio effect implementations in Table \ref{tab:dafxnames}) trained on the VCTK dataset, and compared quantitative performance on the unseen DAPS and MusDB18 datasets. 

We compare the performance of our end-to-end model using both the fixed pretrained audio encoder from Section \ref{sec:audioencoder} as well as training the audio encoder per effect using the SPSA gradients. We also provide a \textit{baseline} experiment, which corresponds to the direct error between the unaffected input and effected ground truth. Similarly to Steinmetz et al. \citep{steinmetz2022style} we recorded the Multi-Resolution STFT (MRSTFT) error (with window sizes 32, 128, 512, 2048, 8192 and 32768) as well as the Perceptual Evaluation of Speech quality (PESQ) as measures of perceptual error. 

\subsubsection{Results and Discussion}

The results in Tables \ref{tab:mrstft} and \ref{tab:pesq} show consistently better performance when using a pretrained encoder, rather than including the audio encoder in the end-to-end training. Further, training was less stable when using uninitialised audio encoder weights, requiring the learning rate to be reduced to $3\mathrm{e}{-5}$ to avoid gradient issues.

However, our approach performs more poorly than the baseline across a number of effects. In particular, style matching modulation and temporal effects such as Ambience and Delay were consistently worse than not applying an effect at all. There may be several factors which contribute to this, such as the encoder not accurately capturing the timbral characteristics of these effects. Analysis of Tables \ref{tab:mrstft} and \ref{tab:pesq} also reveals that model performance is largely consistent across the unseen datasets. 

\subsection{Listening Test}
\label{sec:mushra}

\subsubsection{Experiment Design}

In order to judge perceptual model performance when style matching for \textit{unknown} effect implementations we conduct a listening test, inspired by the Multiple Stimuli with Hidden Reference and Anchor (MUSHRA) design. The test was conducted online using the webMUSHRA framework \citep{schoeffler2018webmushra}. Participants were asked to rate how well a number of conditions match the audio production style of a reference audio source. 

We utilised audio effect implementations from Spotify's Pedalboard library to generate the source audio data, employing four distinct effect types: overdrive/distortion, reverb, delay, and compression/EQ. We transformed audio from the DAPS dataset by applying three different random parameter configurations for each of the effect classes. To achieve style matching, we selected the corresponding effect implementation in the mda-vst library (as outlined in Table \ref{tab:dafxnames}) and the corresponding trained end-to-end model.

The \textit{random baseline}, which acts as a low-quality condition, was generated from a random parameter configuration. The \textit{reference} was also included as a condition with its label hidden. Additionally, for the compression effect, we compared our methodology to the DeepAFx-ST style matching model proposed by Steinmetz et al. \citep{steinmetz2022style}. We did not include this condition for other effects as DeepAFx-ST is only able to style match compression and equalisation. 

\begin{figure*}
    \centering
	\includegraphics[width=\textwidth]{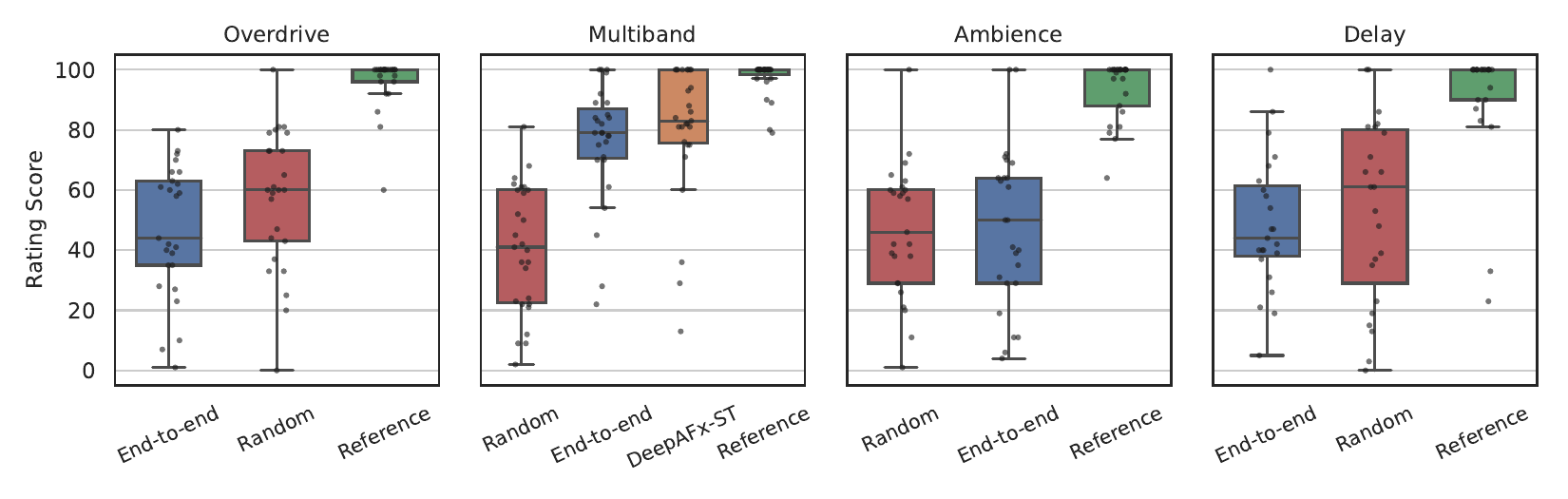}
	\caption{Results of the listening test where our proposed model is labelled \textit{end-to-end}. For the MultiBand effect, the pre-trained DeepAFx-ST model \citep{steinmetz2022style} was also included as a condition.}
    \label{fig:mushra}
\end{figure*}

\subsubsection{Results and Discussion}

From the listening test, we collected responses from 10 participants for our 12 style-matching examples. One submission was disregarded as it rated all samples at 100, leaving nine valid responses for the user evaluation. Therefore, each of the 12 examples garnered a total of 27 ratings for each stimulus. The outcomes of this assessment are illustrated in Figure \ref{fig:mushra}.

Our results show a similar trend to the offline evaluation of our end-to-end network in the previous section. Specifically, our findings indicate that the network was able to better match the style of effects that are more transparent than those with a more obvious acoustic effect. Upon inspection of the audio and parameter configurations predicted by the network, we found that the transformations for Delay and Ambience were subtle in comparison to their given reference. This is in contrast to the random settings of the baseline which, while not matching the exact style of the reference, at least applied a more perceptible effect.

Using an unseen distortion implementation significantly degraded the quality of the style matching for the Overdrive effect. In this case, the effect was subject to the transparency issue seen with other effect classes - i.e, very little drive was applied in all examples.

As seen in Figure \ref{fig:mushra}, the performance of the MultiBand compressor for style matching was significantly better than random settings ($p<1\mathrm{e}{-7}$) and, on average, performed within 6\% of the DeepAFx-ST model \citep{steinmetz2022style}.






\section{Conclusion}

In this work, we have developed a method for audio production style-transfer using non-differentiable audio effects. We trained an audio autoencoder which has been shown to be able to separate digital audio effect classes, and capture parameter changes for a number of effect classes.

This pretrained encoder was then used in a Siamese style-matching network. However, this network was shown to perform poorly with many unseen implementations of effect classes in a user evaluated listening test. Despite this, the approach was shown to work well for a multi-band compressor, whose mean performance was within 6\% of a state-of-the-art approach using auto-differentiation.

In future work, we suggest revisiting the encoder network and examining the possible advantages of conditioning on the latent space and employing distinct encoders for effect classes. One approach for the former would be to use perceptually-regulated variational timbre spaces as proposed by Esling et al. \citep{esling2018bridging}. The latter can be achieved by assigning an encoder for each umbrella effect class (e.g. temporal effects, modulation effects, etc.). This may encourage the encoder to learn features that are more specific to that particular class of effect.


\bibliographystyle{jaes}

\bibliography{refs}

\end{document}